\documentclass[conference]{IEEEtran}
\IEEEoverridecommandlockouts

\usepackage{cite}
\usepackage{amsmath,amssymb,amsfonts}
\usepackage{algorithmic}
\usepackage{graphicx}
\usepackage{textcomp}
\usepackage{comment}
\usepackage{xcolor}
\usepackage{subcaption}
\usepackage[colorlinks, linkcolor=blue, citecolor=blue, urlcolor=blue]{hyperref}
\usepackage{float}
\usepackage{fancyhdr}
\usepackage{multirow}
\usepackage{enumerate}
\usepackage{balance}

\def\BibTeX{{\rm B\kern-.05em{\sc i\kern-.025em b}\kern-.08em
    T\kern-.1667em\lower.7ex\hbox{E}\kern-.125emX}}

\fancypagestyle{firstpage}
{
    \fancyhead[R]{} 
    \fancyhead[L]{2024 27th International Conference on Computer and Information Technology (ICCIT),\\ 20-22 December 2024, Cox’s Bazar, Bangladesh}
    \fancyfoot[C]{} 
    \fancyfoot[L]{979-8-3315-1909-4/24/\$31.00 ©2024 IEEE}
    \setlength{\footskip}{25pt} 
}

\begin{document}

\makeatletter
    \newcommand{\linebreakand}{%
      \end{@IEEEauthorhalign}
      \hfill\mbox{}\par
      \mbox{}\hfill\begin{@IEEEauthorhalign}
    }
\makeatother

\title{Explainable AI-Enhanced Deep Learning for Pumpkin Leaf Disease Detection: A Comparative Analysis of CNN Architectures\\
}

\author{\IEEEauthorblockN{ Md. Arafat Alam Khandaker\IEEEauthorrefmark{1}, Ziyan Shirin Raha\IEEEauthorrefmark{1}, Shifat Islam\IEEEauthorrefmark{2}, Tashreef Muhammad\IEEEauthorrefmark{2}
}
\IEEEauthorblockA{\IEEEauthorrefmark{1}Department of Computer Science and Engineering, Ahsanullah University of Science and Technology, Dhaka, Bangladesh}
\IEEEauthorblockA{\IEEEauthorrefmark{2}Department of Computer Science and Engineering, Bangladesh University of Engineering and Technology, Dhaka, Bangladesh}
\IEEEauthorblockA{\IEEEauthorrefmark{3}Department of Computer Science and Engineering, Southeast University, Dhaka, Bangladesh}

aarafatalam18@gmail.com, ziyanraha@gmail.com, 
shifat.islam.buet@gmail.com, tashreef.muhammad@seu.edu.bd   
}


\maketitle
\thispagestyle{firstpage}
\begin{abstract}
Pumpkin leaf diseases are significant threats to agricultural productivity, requiring a timely and precise diagnosis for effective management. Traditional identification methods are laborious and susceptible to human error, emphasizing the necessity for automated solutions. This study employs on the ``Pumpkin Leaf Disease Dataset'', that comprises of 2,000 high-resolution images separated into five categories. Downy mildew, powdery mildew, mosaic disease, bacterial leaf spot, and healthy leaves. The dataset was rigorously assembled from several agricultural fields to ensure a strong representation for model training. We explored many proficient deep learning architectures, including DenseNet201, DenseNet121, DenseNet169, Xception, ResNet50, ResNet101 and InceptionResNetV2, and observed that ResNet50 performed most effectively, with an accuracy of 90.5\% and comparable precision, recall, and F1-Score. We used Explainable AI (XAI) approaches like Grad-CAM, Grad-CAM++, Score-CAM, and Layer-CAM to provide meaningful representations of model decision-making processes, which improved understanding and trust in automated disease diagnostics. These findings demonstrate ResNet50's potential to revolutionize pumpkin leaf disease detection, allowing for earlier and more accurate treatments.

\end{abstract}

\begin{IEEEkeywords}
Pumpkin leaf detection, Deep Learning, CNN Architecture, Explainable AI
\end{IEEEkeywords}
\section{Introduction}
\label{sec:introduction}
Pumpkin leaf diseases including downy mildew, powdery mildew, mosaic disease, and bacterial leaf spot are significant threats to the production of agriculture, culminating with substantial crop losses. These diseases not only diminish yields but also have an impact on crop quality, making effective disease control vital for farmers. Early and precise detection of these diseases is critical for reducing their impact, but traditional procedures based on manual inspection are frequently time-consuming, labour-intensive, and prone to human error. As a result, these problems highlight the importance of efficient, automated, and accurate diagnostic approaches that can be used at scale in agricultural contexts. Deep learning has transformed areas like healthcare and automotive by automating complicated operations \cite{c1}, but its applicability in agriculture is still underexplored. One of the most significant challenges to its widespread adoption in agriculture is a lack of transparency in how AI models make verdicts, which is critical for creating assurance and ensuring practical utility in real-world circumstances.

Explainable AI (XAI) \cite{c2} provides an appealing approach for making AI models more transparent and intelligible to users. XAI approaches such as visual heatmaps assist in emphasizing the essential elements of images that influence a model's prediction, providing insights into decision-making processes. This is especially crucial in agriculture, where openness and confidence in AI-driven decision-making may help farmers and agricultural experts make sound decisions based on trustworthy AI insights. The inclusion of AI increases its potential for clinical diagnosis and research \cite{c3}. This formidable combination has the potential to broaden its applications across a variety of fields, including potato disease detection \cite{c4}, detection of plant leaf disease \cite{c5}, plant disease recognition \cite{c6} and automatic detection of fruit disease \cite{c7}.

Standard convolutional neural networks (CNN) models, that include DenseNet \cite{c8}, ResNet \cite{c14}, Xception\cite{c15}, and InceptionResNetV2 \cite{c16}, perform well in detecting plant diseases, but have poor interpretability. This study aims to address this issue by using cutting-edge XAI approaches such as Grad-CAM \cite{c17}, Grad-CAM++ \cite{c18}, Score-CAM \cite{c19}, and Layer-CAM \cite{c20} to produce visual explanations of model predictions. These techniques not only improve transparency but also make agricultural applications more practical by highlighting the areas of leaf images that are most relevant to disease classification. The research paper's contributions are summarized below: 
\begin{itemize}
    \item We evaluated seven pre-trained convolutional neural network (CNN) models (ResNet50, ResNet101, DenseNet121, DenseNet169, DenseNet201, Xception) and InceptionResNetV2) for automated pumpkin leaf classification.
    \item Explainable AI techniques including GradCam, GradCam++, ScoreCam, and LayerCam are being utilized to improve transparency.  
    \item This research expands the gap in Explainable Artificial Intelligence (XAI) technique for the pumpkin classification challenge by being the first to use XAI techniques in this setting and providing novel insights into model interpretability.
\end{itemize}
Our study improves the field of AI in agriculture by incorporating XAI into the diagnostic procedure, which leads to high-performing and interpretable models that are essential to effective disease management decision-making.\\
In this study, we have already explained the introduction in Section \ref{sec:introduction}. A thorough related works review will follow it in Section \ref{sec:Related Works}. Then in different subsections, we discuss the conducted work’s methodology in Section \ref{sec:methodology}. Then we describe our results in Section \ref{sec: result anslysis}. Afterwards that, we are presented our limitations and future work in section \ref{sec:limitation} followed by a conclusion in Section \ref{sec:conclusion} to summarize our conducted study.

\section{Related Works}
\label{sec:Related Works}
We have reviewed several studies on pumpkin disease detection. However, none of them have incorporated Explainable AI (XAI) techniques into their methodologies. This section offers a concise review of the existing research consisting of Machine Learning and Deep Learning Models. It highlights the relevant methods and their respective contributions to the field.
Mosaddek Ali et al. \cite{c3} demonstrated the effectiveness of traditional machine learning models, such as SVM, KNN, Decision Tree, and Naive Bayes, with CNNs for detecting pumpkin leaf diseases. in agriculture, employing a dataset of 2000 pumpkin leaf pictures to reach a high validation accuracy of 87.67\%. This work demonstrates CNN's capacity to extract significant features from picture data, which is critical for accurate disease identification. Moreover, Yousef Methkal et al. \cite{c11} developed an ACO-CNN model that improves classic CNN architecture with Ant Colony Optimization. This approach not only obtained an outstanding 99.98\% accuracy, but also maintained good precision, recall, and F1-scores, demonstrating the advantages of hybrid models in complicated classification problems. Furthermore, Naeem et al. \cite{c6} developed DeepPlantNet, a 28-layer deep learning network designed for plant disease classification. This model obtained accuracies of 98.49\% and 99.85\% for its eight- and three-class classification tasks, respectively, demonstrating advances in deep learning for agriculture. highlighting the adaptability of CNNs.

However, Asad et al. \cite{c7} utilized a CNN model to identify and categorize common citrus diseases, achieving a test accuracy of 94.55\%. This indicates CNNs' broad usefulness, not just in leaf disease detection, but also in citrus disease management, which presents distinct obstacles. Besides, Sk Mahmudul et al. \cite{c9} investigated different CNN architectures on the PlantVillage dataset and found that EfficientNetB0 was the most efficient. This demonstrates the efficacy of advanced CNN models in plant disease detection and how architectural differences might affect detection accuracy. 
Additionally, Eman Abdullah et al. \cite{c5} leveraged the PlantVillage dataset to implement the YOLOv4 algorithm for real-time disease identification and classification. The model's excellent accuracy of 99.99\% highlights its versatility in operating contexts where speed and accuracy are critical.

Besides, Mobeen et al. \cite{c10} proposed an efficient CNN-based technique adapted for mobile devices, utilizing MobileNetV3Large. The study focused on class imbalance and resource constraints, demonstrating that stepwise transfer learning can successfully improve model performance while remaining resource-economical.
Whereas, Touhidul et al. \cite{c13} compared the performance and computational efficiency of a bespoke CNN, LDDTA, to nine pre-trained models. LDDTA's competitive performance shows that simpler models can nevertheless achieve excellent accuracy, making them appropriate for applications with limited computational resources. Therefore, Alam et al. \cite{c12} deployed  CycleGAN and Pix2Pix models are implemented to generate synthetic potato disease images, increasing dataset diversity and classification and segmentation efficiency. CycleGAN outperformed Pix2Pix, with better Inception Scores and lower Fréchet Inception Distances. Significant breakthroughs include the novel use of GANs for data augmentation, the use of Explainable AI approaches (GradCAM, GradCAM++, ScoreCAM) to improve model interpretability, and the incorporation of Detectron2 for accurate disease segmentation. These developments greatly improve potato disease detection in agriculture.

 \section{Methodology }
 \label{sec:methodology}

\begin{figure*}[h]
    \centering
    \begin{subfigure}[h]{0.19\textwidth}
        \includegraphics[width=\textwidth]{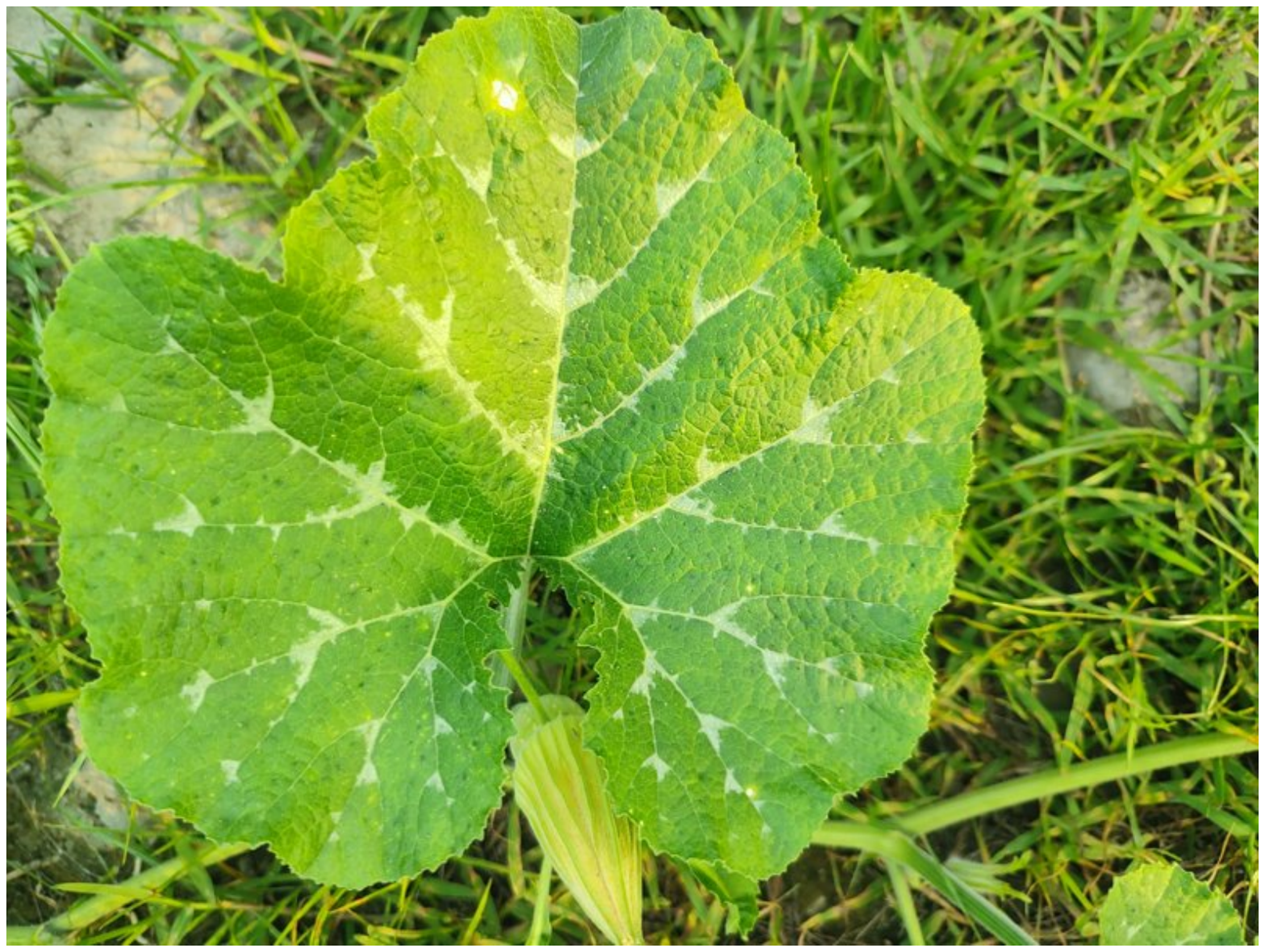}
        \caption{Bacterial Leaf Spot}
        \label{fig:sub1}
    \end{subfigure}
    \begin{subfigure}[h]{0.19\textwidth}
        \includegraphics[width=\textwidth]{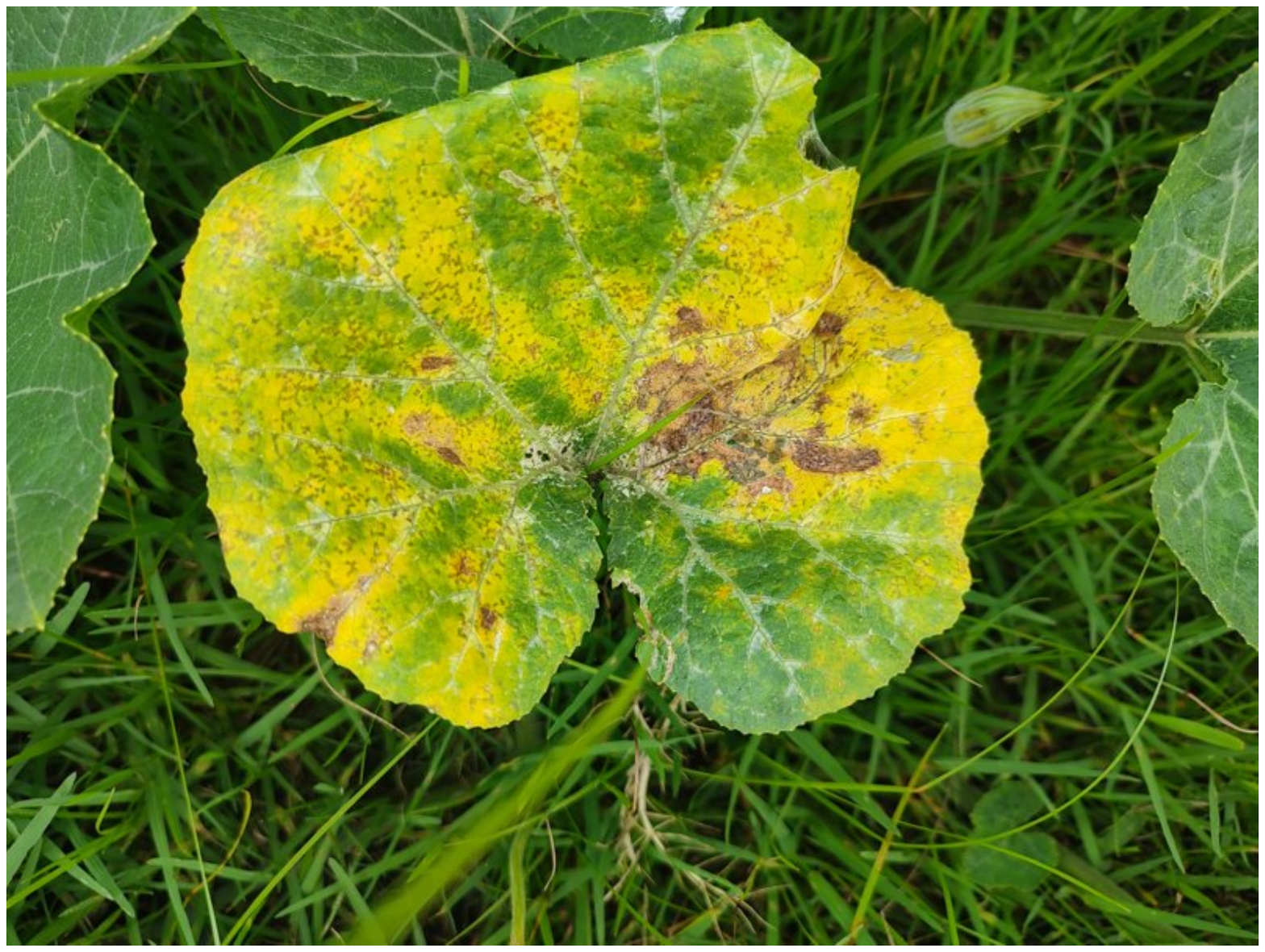}
        \caption{Downy Mildew}
        \label{fig:sub2}
    \end{subfigure}
      \begin{subfigure}[h]{0.19\textwidth}
        \includegraphics[width=\textwidth]{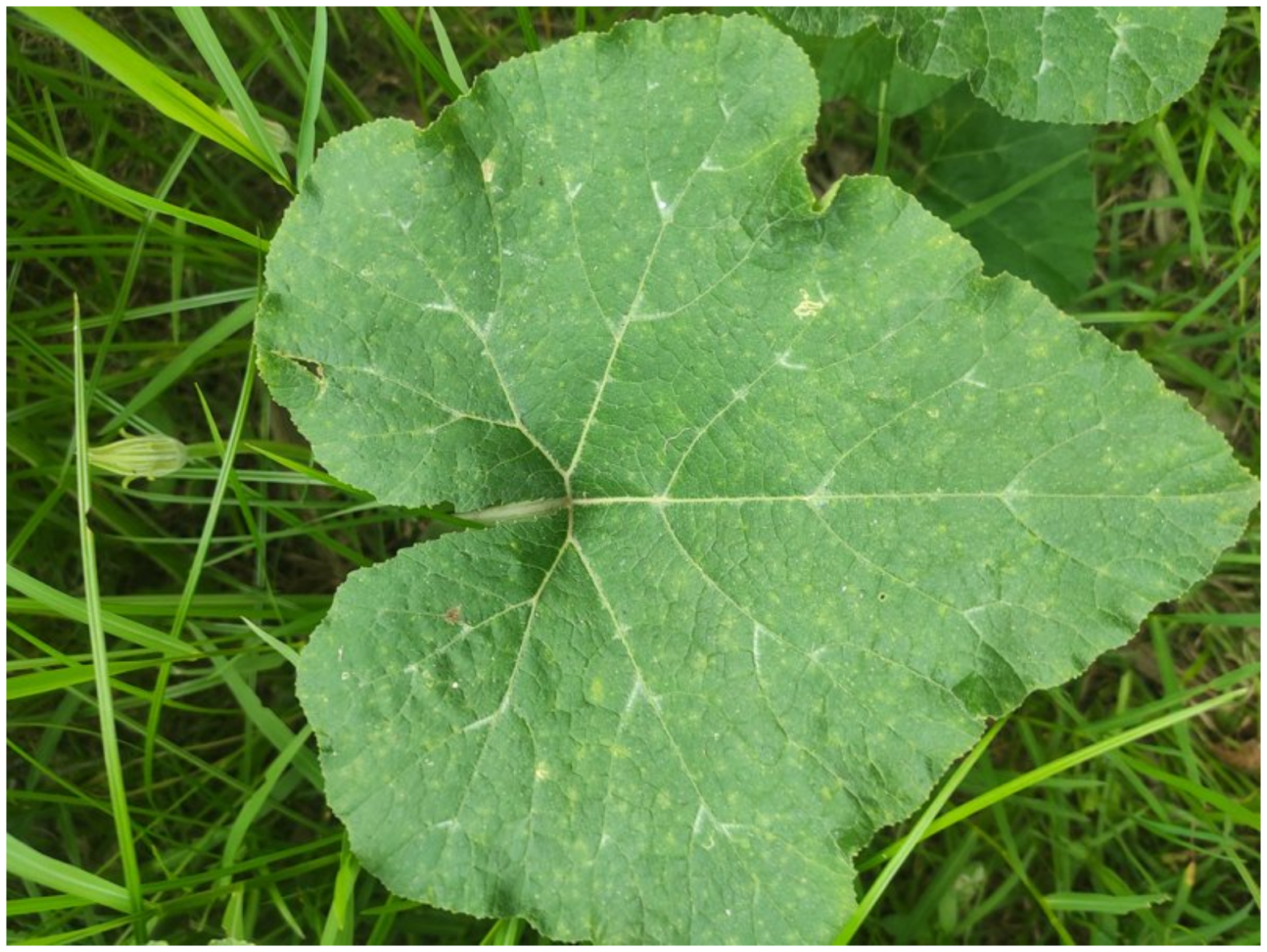}
        \caption{Healthy Leaf}
        \label{fig:sub3}
    \end{subfigure}
      \begin{subfigure}[h]{0.19\textwidth}
        \includegraphics[width=\textwidth]{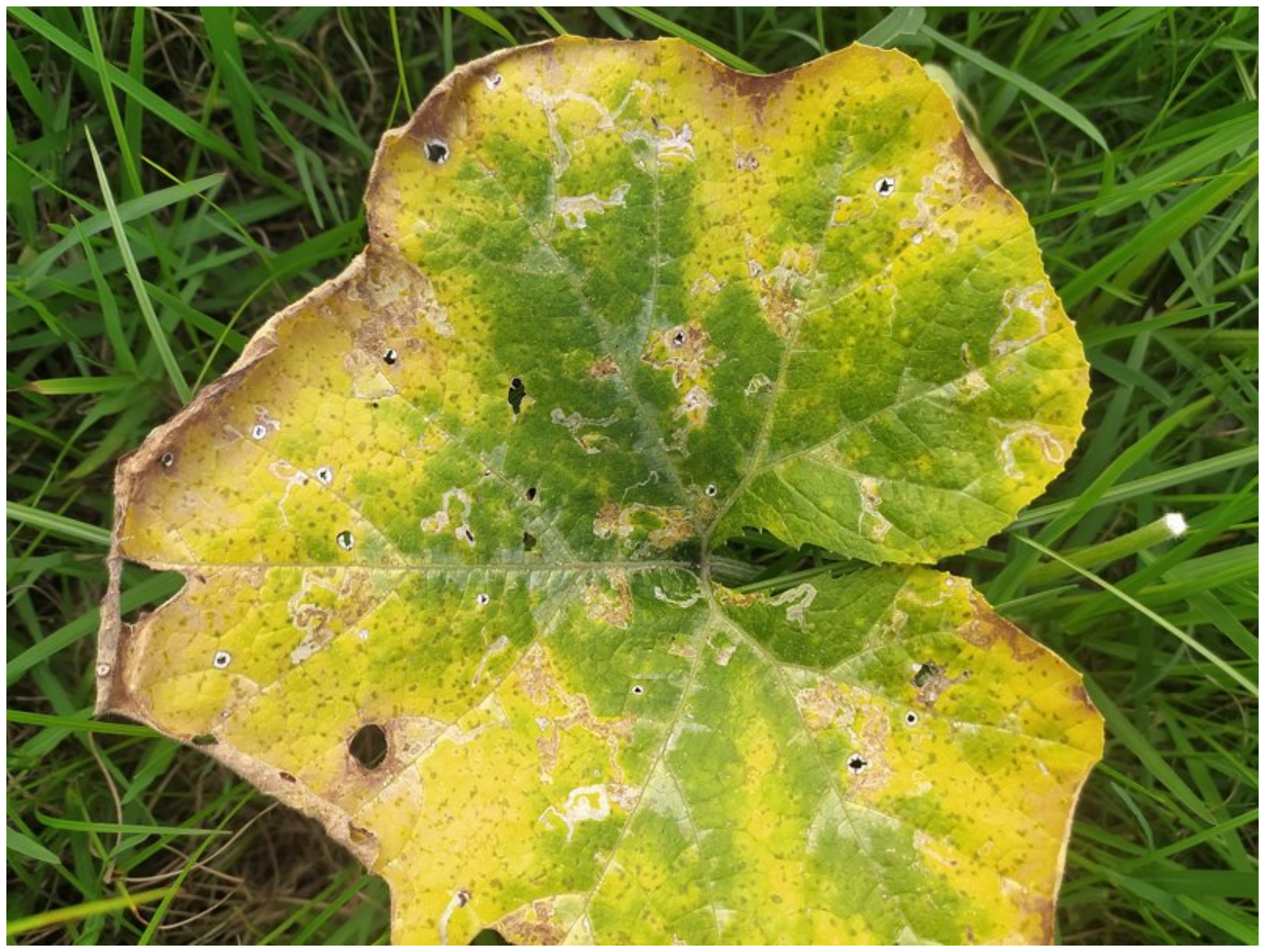}
        \caption{Mosaic Disease}
        \label{fig:sub4}
    \end{subfigure}
    \begin{subfigure}[h]{0.19\textwidth}
        \includegraphics[width=\textwidth]{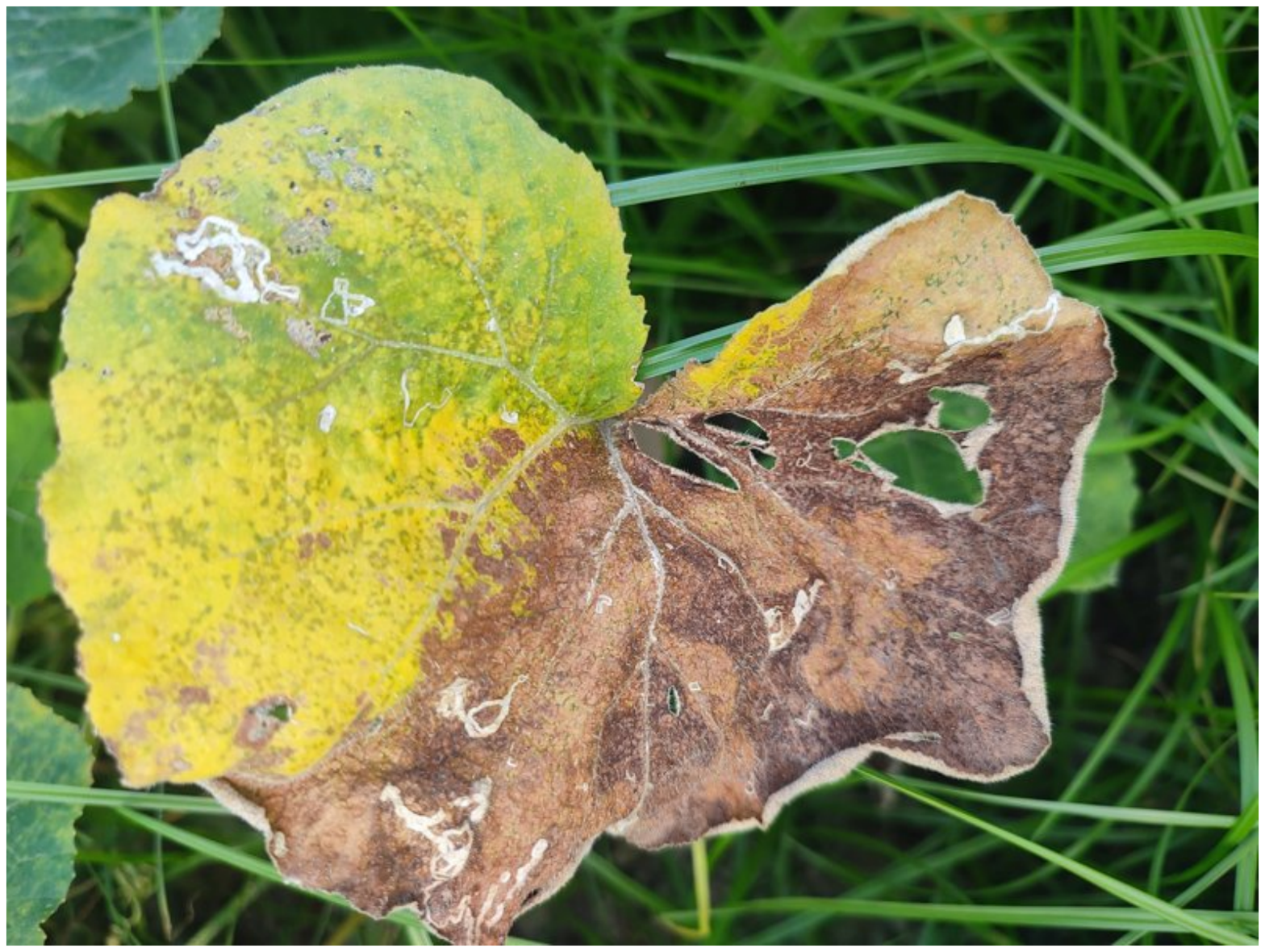}
        \caption{Powdery Mildew}
        \label{fig:sub5}
    \end{subfigure}
    \caption{Displays Representation Samples from the Dataset Showcasing Various Categories of Pumpkin Leaves}
    \label{fig:dataset_example}
\end{figure*}

\begin{figure*}[t]
\centering
\includegraphics[width=1\textwidth]{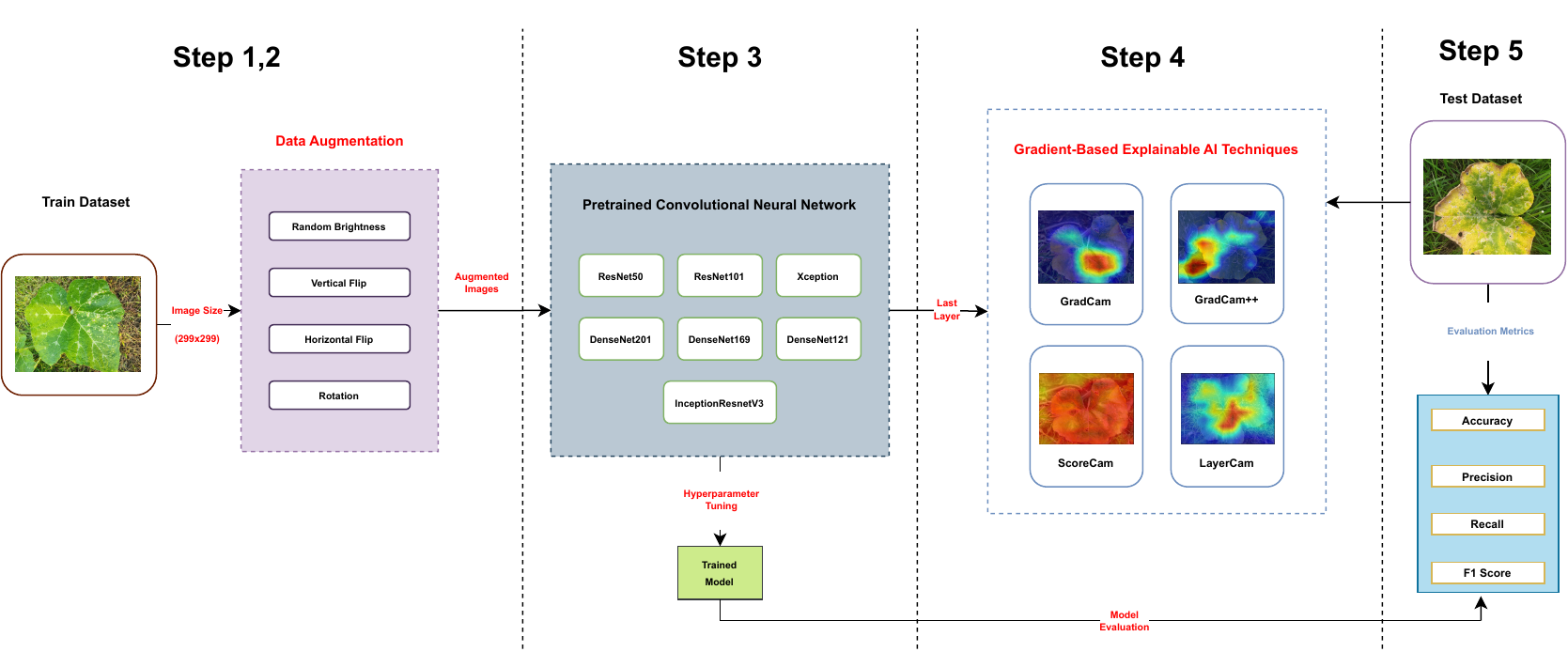}
\caption{Implied Methodology for Classifying Pumpkin Leaf Using Explainable AI Techniques}
\label{fig:methodology}
\end{figure*}

\subsection{Dataset}
`` The Pumpkin Leaf Diseases Dataset ”  \cite{c21} is an extensive resource that includes 2,000 high-resolution images of healthy and diseased pumpkin leaves organized into categories including Downy Mildew, Powdery Mildew, Mosaic Disease, Bacterial Leaf Spot, and Healthy Leaves. Each image is labeled dependent on whether the pumpkin leaf is diseased or healthy. The images are evenly distributed throughout these categories, yielding a balanced dataset for disease classification and analysis. It is obtained from Mendeley, reflecting real-world agricultural settings, including a variety of environmental factors, leaf sizes, and disease severity levels. It is precisely intended to help plant pathology and machine learning research, with applications in disease categorization, diagnosis, and automated detection systems for precision agriculture. This dataset makes it easier to create interpretable AI models by providing prospective insights via Explainable Artificial Intelligence (XAI) methodologies, all while encouraging innovation in sustainable agricultural management and educational practices. Its significance is in confronting agricultural challenges through enabling effective disease monitoring, optimizing pesticide use, and enhancing crop quality, making it a must-have tool for developing plant disease research and pragmatic agricultural solutions. 

\subsection{ Proposed Methodology}
To compare the effectiveness of different models in evaluating pumpkin leaf disease images, we developed a methodology, as defined in  fig. \ref{fig:methodology}  of our study. This methodology incorporates numerous phases of data preprocessing, model training, and evaluation to ensure an extensive investigation.\\

\textbf{Step 1)  Input Image:}
 To ensure the reliability of the various models and techniques, basic processing steps were conducted to each image, commencing with an adjustment to a standard 256 × 256 pixel size.

\textbf{Step 2) Image Preprocessing Technique:} Normalizing pixel values on a scale of 0 to 1 assisted in facilitating a consistent flow of information and increased the rate of closure when training.

\textbf{Data Augmentation:} The model's ability to generalize across numerous orientations was enhanced by using random rotations to introduce diversity in image orientation. In addition, mirrored images were simulated using horizontal flipping, increasing dataset variation while decreasing the likelihood of overfitting.

\textbf{Cropping and Focus Enhancement:} Images were resized consciously to eliminate inefficient background components and highlight areas of interest. This allowed us to concentrate on important features while minimizing computational overhead.

\textbf{Brightness Adjustment:} Image brightness varies were applied to reduce the impact of lighting alterations throughout the set of images, resulting in uniform exposure levels.

\textbf{Contrast Enhancement:} These methods were used to improve image contrast, which is significant for healthcare imaging since it allows for accurate classification of small attributes.

\textbf{Step 3) Model Selection and Training:} For this investigation, we evaluated seven Convolutional Neural Network (CNN) architectures known for their picture classification performance: Xception, DenseNet201, DenseNet121, DenseNet169, InceptionResnetV2, ResNet50, and ResNet101. The preprocessed dataset was used during training to ensure stability and consistency. Hyperparameter optimization was employed to fine-tune each model for the particular aim of diagnosing pumpkin leaf disease. 

\textbf{Step 4) Application of Explainable AI (XAI) Techniques:} To enhance understanding of the model's decisions, we deployed Explainable AI approaches to the CNNs' final layers. GradCAM, GradCAM++, ScoreCAM, and LayerCAM were used to create heatmaps that highlight the most important parts of the photos for each class, allowing us to identify which image aspects had the greatest influence on the model's decision-making process. These methodologies enabled class-specific attention mapping and multi-layer activation visualization, resulting in more detailed insights into model performance.

\textbf{Step 5) Performance Evaluation:} We used a variety of metrics to evaluate each model's capacity to spot leaf disease images. This thorough review enabled us to assess and compare the performance of numerous models, allowing us to select the best one for our categorization task. Metrics comprising as accuracy, precision, recall, and F1-score presented provide an exhaustive overview of each model's performance in multiple aspects of the classification task. Furthermore, among emphasizing each model's advantages and disadvantages in managing different disease categories, these measures provide a fair assessment.\\

\section{Result Analysis}
\label{sec: result anslysis}

\subsection{Experimental setup}

Experiments were carried out using the Kaggle Notebooks platform and the Tesla P100 GPU computing engine. The dataset is separated into train, validation, and test instances. The training set takes up 80\% of the dataset, while the remaining 20\% is split into two sets: validation (10\%) and test (10\%).

\begin{table}[htbp]
\caption{Training Parameters of Proposed Models}
\begin{center}
\setlength{\tabcolsep}{25pt}
\renewcommand{\arraystretch}{1.8}
\begin{tabular}{|c|c|}
\hline
\textbf{Training Parameters} & \textbf{Value} \\
\hline
No. of epochs & 30,50,100 \\
\hline
Optimization algorithm & Adam\\
\hline
Input image size & 299x299\\
\hline
Batch size & 6, 8, 10, 12\\
\hline
Learning rate & 1e-3, 1e-5\\
\hline
\end{tabular}
\label{tab1}
\end{center}
\end{table}

\subsection{Evaluation}

The evaluation is conducted by considering performance metrics such as accuracy, precision, recall and F1-score  which are presented in TABLE \ref{tab2} 

\begin{table}[htbp]
\caption{Assessing the Performance of Diverse Pre-trained Models}
\begin{center}
\setlength{\tabcolsep}{3.5pt}
\renewcommand{\arraystretch}{2.3}
\begin{tabular}{|c|c|c|c|c|c|}
\hline
\textbf{Architecture} & \textbf{Accuracy} & \textbf{Precision} & \textbf{Recall} & \textbf{F1-Score}\\
\hline
DenseNet201 & 0.845 & 0.843 & 0.845 & 0.8424\\
\hline
DenseNet121 & 0.841 & 0.8388 & 0.841 & 0.836\\
\hline
DenseNet169 & 0.862 & 0.8609 & 0.862 & 0.8586 \\
\hline
\textbf{ResNet50} & \textbf {0.905} & \textbf{0.9044} & \textbf{0.905} & \textbf{0.9041}\\
\hline
ResNet101 & 0.875 & 0.8604 & 0.877 & 0.8588\\
\hline
InceptionResnetV2 & 0.815 & 0.816 & 0.814 & 0.813\\
\hline
Xception & 0.885 & 0.823 & 0.825 & 0.820 \\
\hline
\end{tabular}
\label{tab2}
\end{center}
\end{table}

ResNet50 exceeded all other deep learning models in the pumpkin leaf disease classification assessment, with the highest accuracy (0.905), precision (0.9044), recall (0.905), and F1-score (0.9041). These findings were achieved with a batch size of 10, 50 epochs, and a learning rate of 1e-5, which was revealed to be the most effective combination. ResNet50 outperforms its deeper version, ResNet101 (accuracy 0.875, F1-score 0.8588), demonstrating that ResNet50's balanced architecture, with remaining connections that avoid gradient vanishing, allows for improved generalization on this dataset. DenseNet169 outperformed the rest of the DenseNet models, with an F1-score of 0.8586 and an accuracy of 0.862. DenseNet201 and DenseNet121 followed modestly. Xception (accuracy 0.885, F1-score 0.820) scored better than InceptionResNetV2 (accuracy 0.815, F1-score 0.813), although ResNet and DenseNet architectures combine both.
\begin{figure*}[t]
    \centering
    
    \begin{subfigure}[b]{0.24\textwidth}
        \fcolorbox{lightgray}{white}{\includegraphics[width=\textwidth]{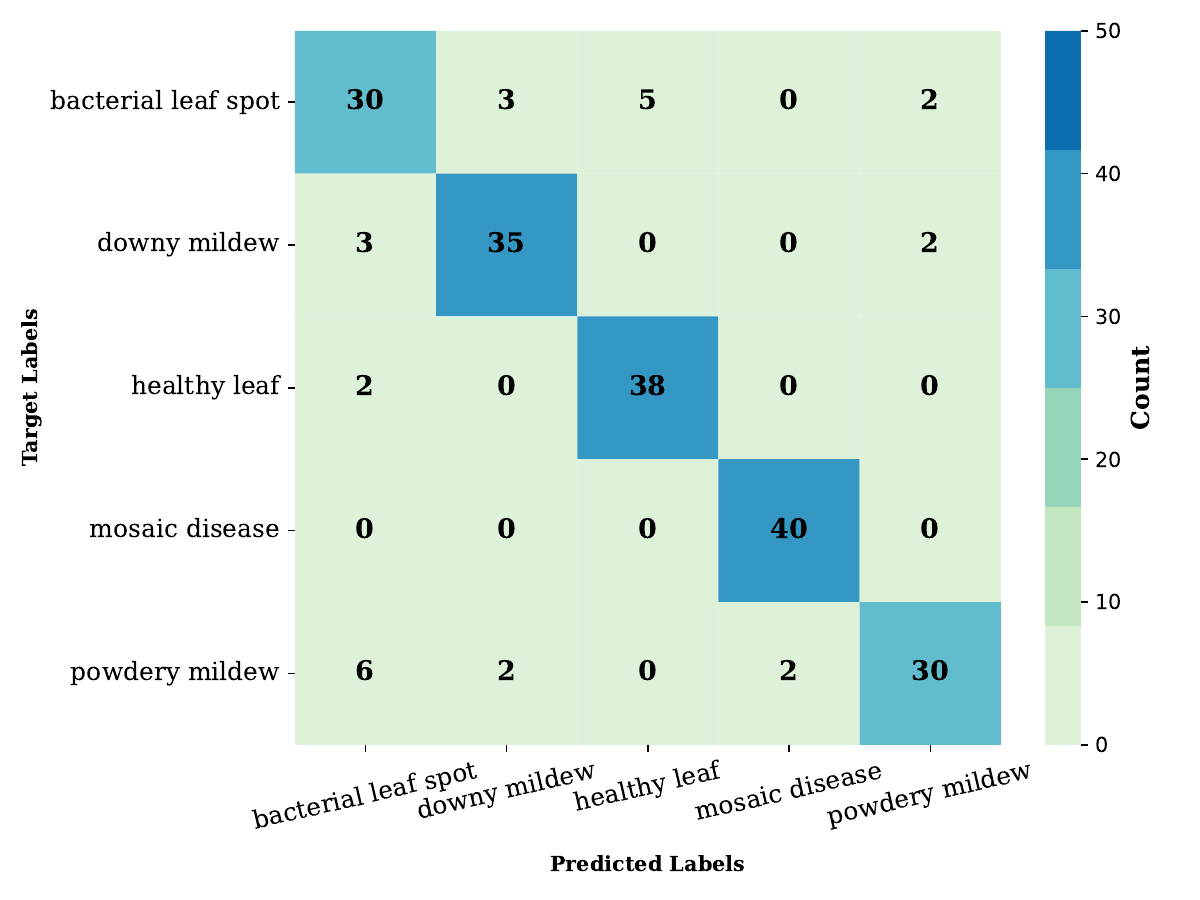}}
        \caption{DenseNet201}
        \label{fig:sub1}
    \end{subfigure}
    \hfill
    \begin{subfigure}[b]{0.24\textwidth}
        \fcolorbox{lightgray}{white}{\includegraphics[width=\textwidth]{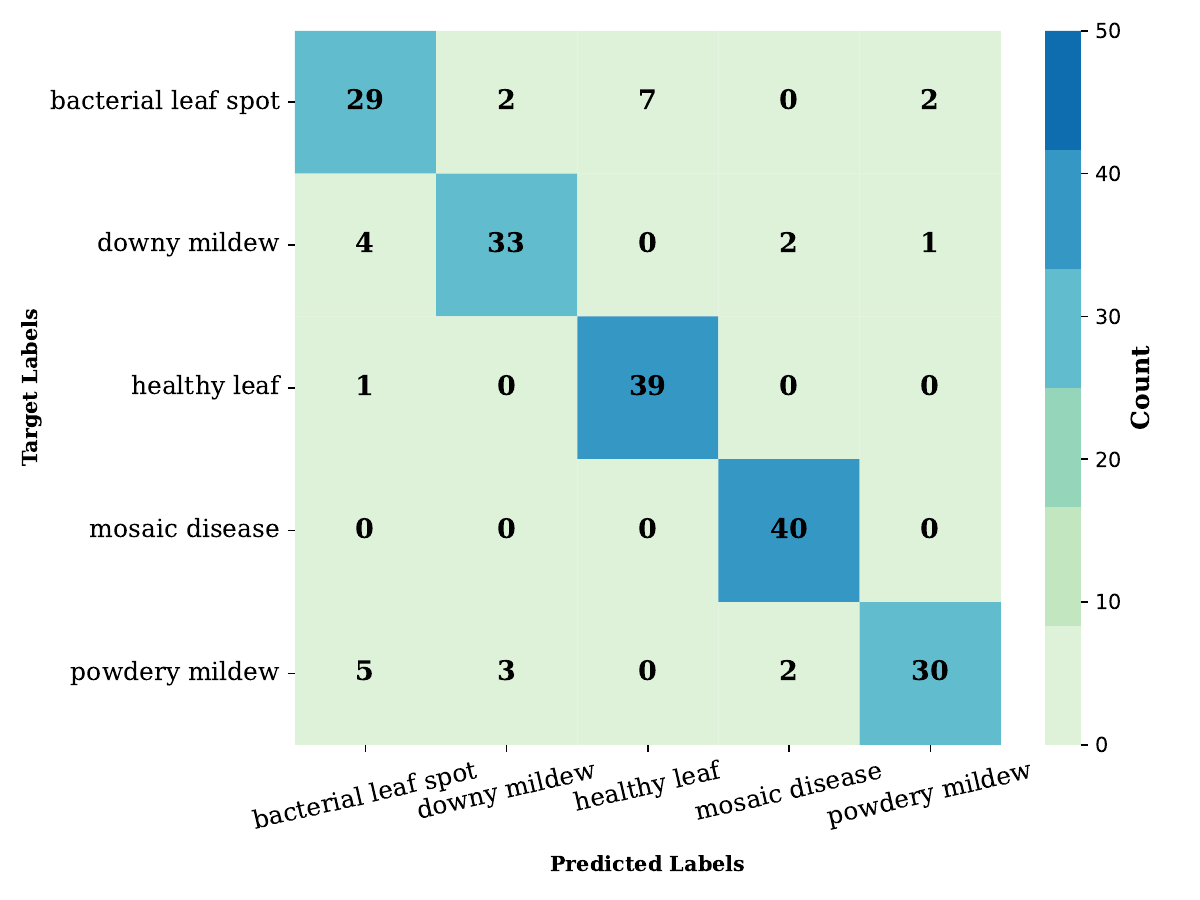}}
        \caption{DenseNet169}
        \label{fig:sub2}
    \end{subfigure}
    \hfill
    \begin{subfigure}[b]{0.24\textwidth}
        \fcolorbox{lightgray}{white}{\includegraphics[width=\textwidth]{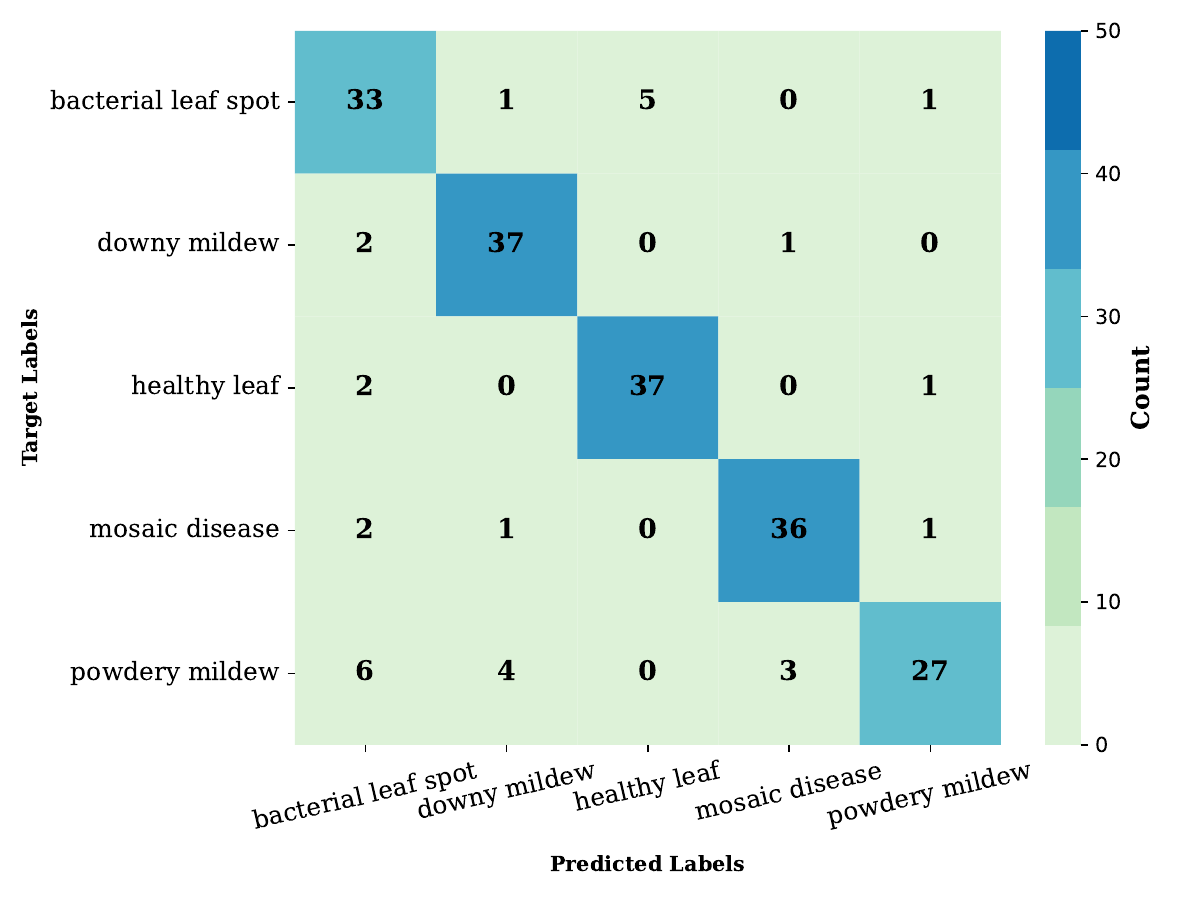}}
        \caption{DenseNet121}
        \label{fig:sub3}
    \end{subfigure}
    \hfill
    \begin{subfigure}[b]{0.24\textwidth}
        \fcolorbox{lightgray}{white}{\includegraphics[width=\textwidth]{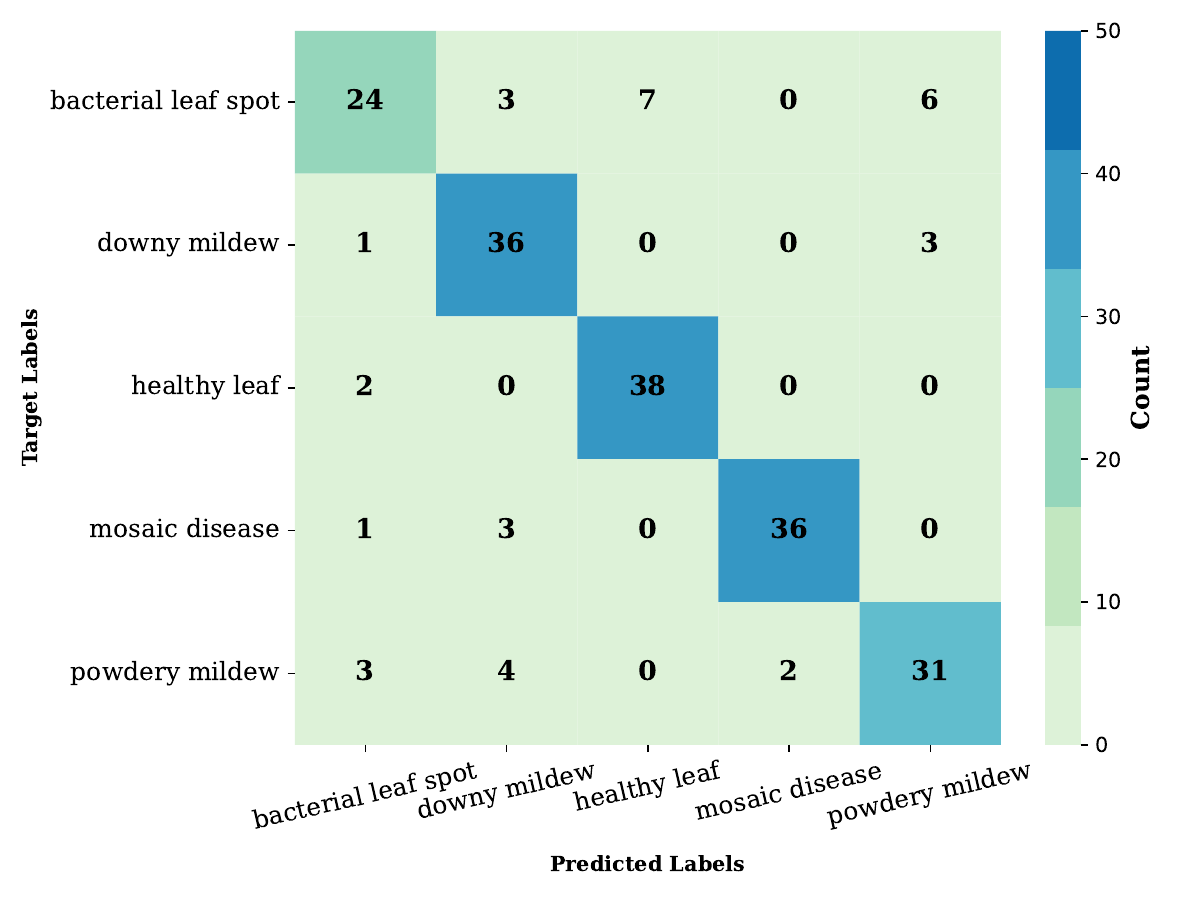}}
        \caption{ResNet101}
        \label{fig:sub4}
    \end{subfigure}
    \hfill
    \begin{subfigure}[b]{0.24\textwidth}
        \fcolorbox{lightgray}{white}{\includegraphics[width=\textwidth]{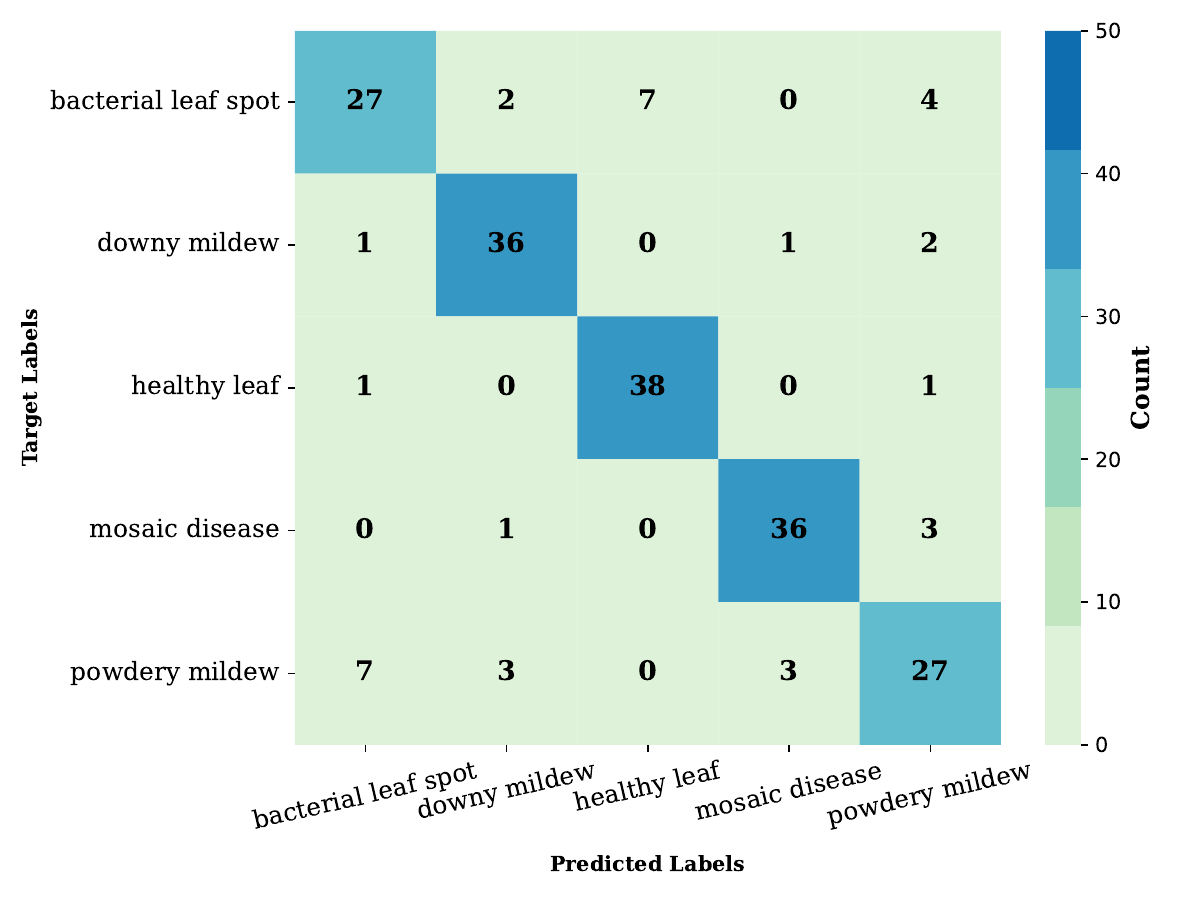}}
        \caption{Xception}
        \label{fig:sub5}
    \end{subfigure}
    \hfill
    \begin{subfigure}[b]{0.24\textwidth}
        \fcolorbox{lightgray}{white}{\includegraphics[width=\textwidth]{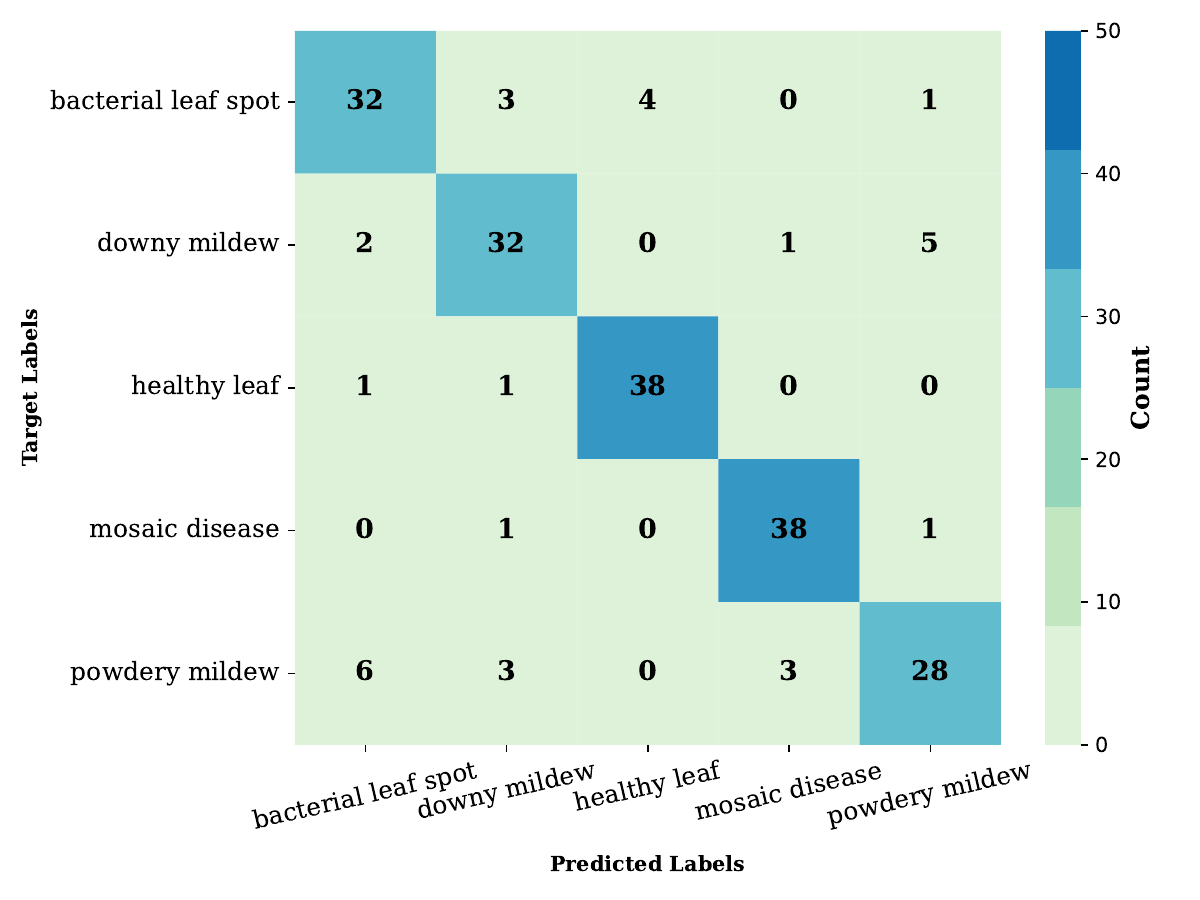}}
        \caption{InceptionResNetV2}
        \label{fig:sub6}
    \end{subfigure}
    \hfill
    \begin{subfigure}[b]{0.24\textwidth}
        \fcolorbox{lightgray}{white}{\includegraphics[width=\textwidth]{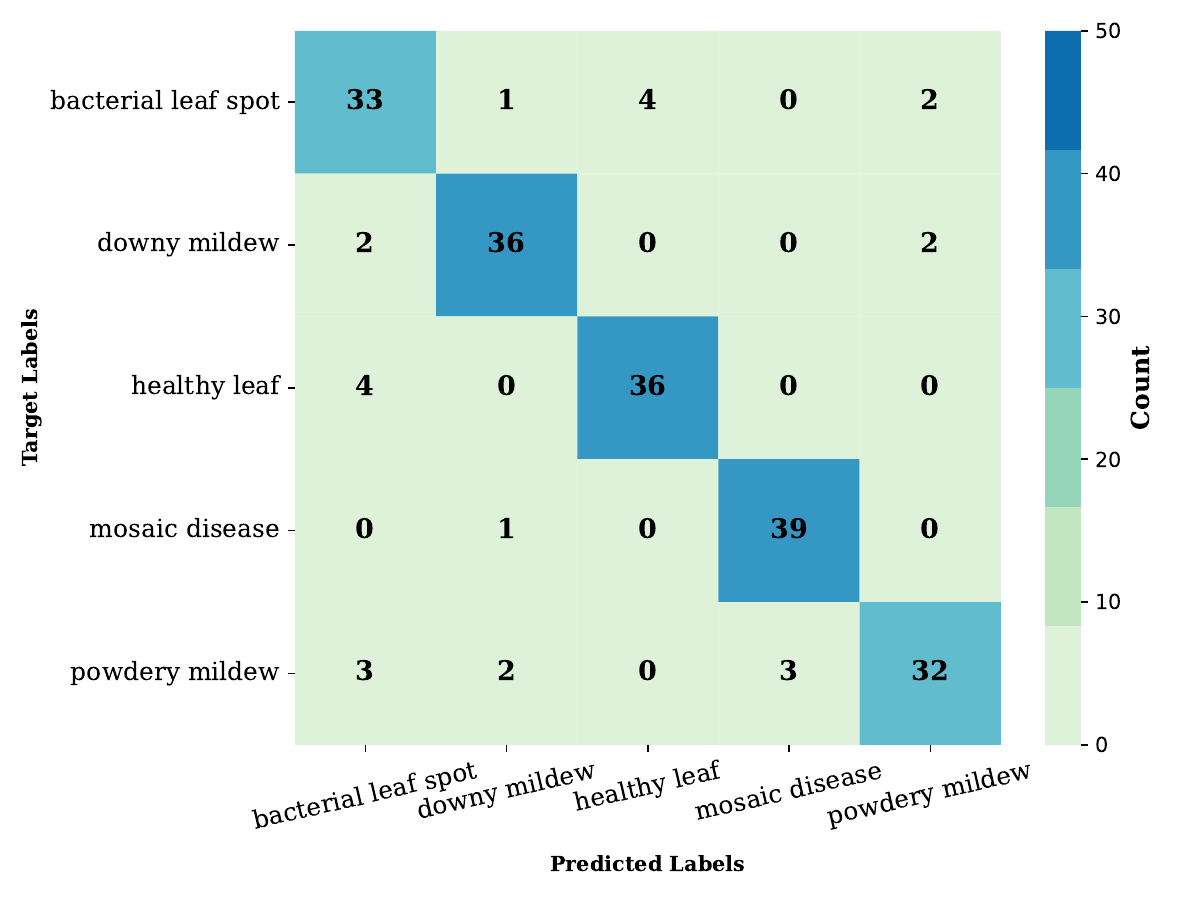}}
        \caption{ResNet50}
        \label{fig:sub7}
    \end{subfigure}
    \hfill
    \caption{Confusion Matrix of Pre-Trained CNNs for Pumpkin Leaf Classification}
    \label{fig:confusionmatrix}
\end{figure*}
 \begin{figure*}[h]
    \centering
    \begin{subfigure}[b]{0.24\textwidth}
        \includegraphics[width=\textwidth]{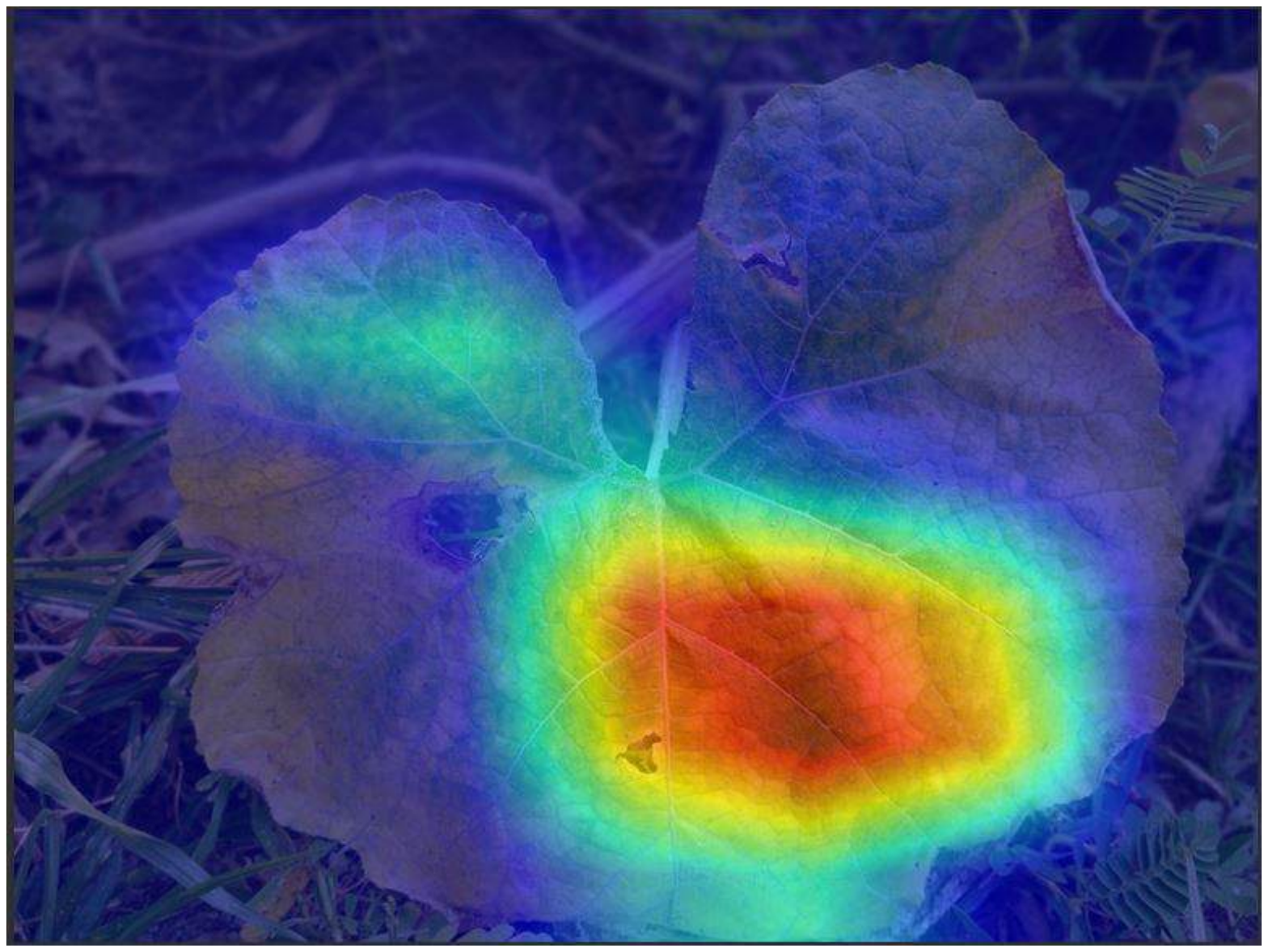}
        \caption{GradCAM}
        \label{fig:sub1}
    \end{subfigure}
    \hfill
    \begin{subfigure}[b]{0.24\textwidth}
        \includegraphics[width=\textwidth]{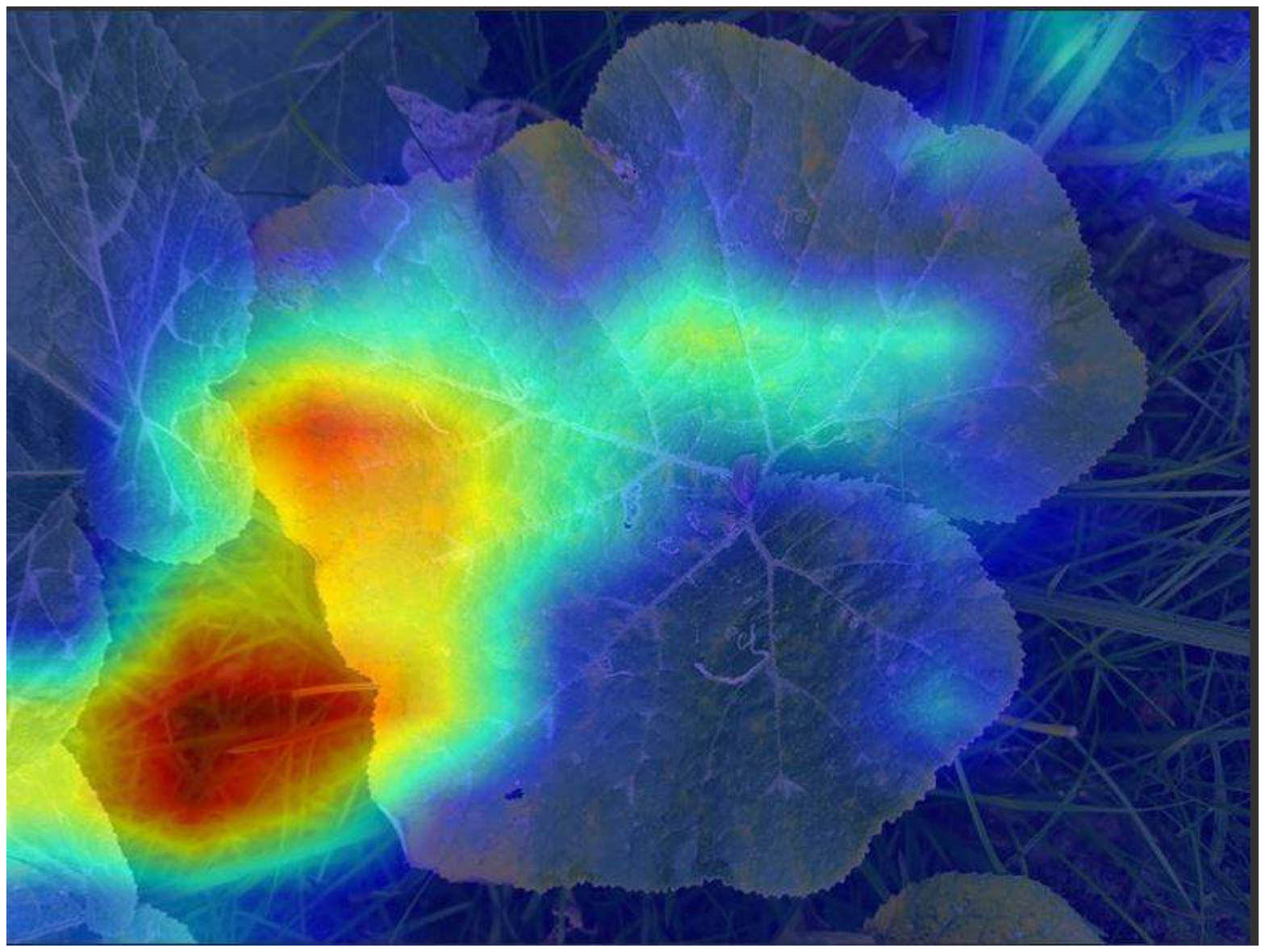}
        \caption{GradCAM++}
        \label{fig:sub2}
    \end{subfigure}
    \hfill
    \begin{subfigure}[b]{0.24\textwidth}
        \includegraphics[width=\textwidth]{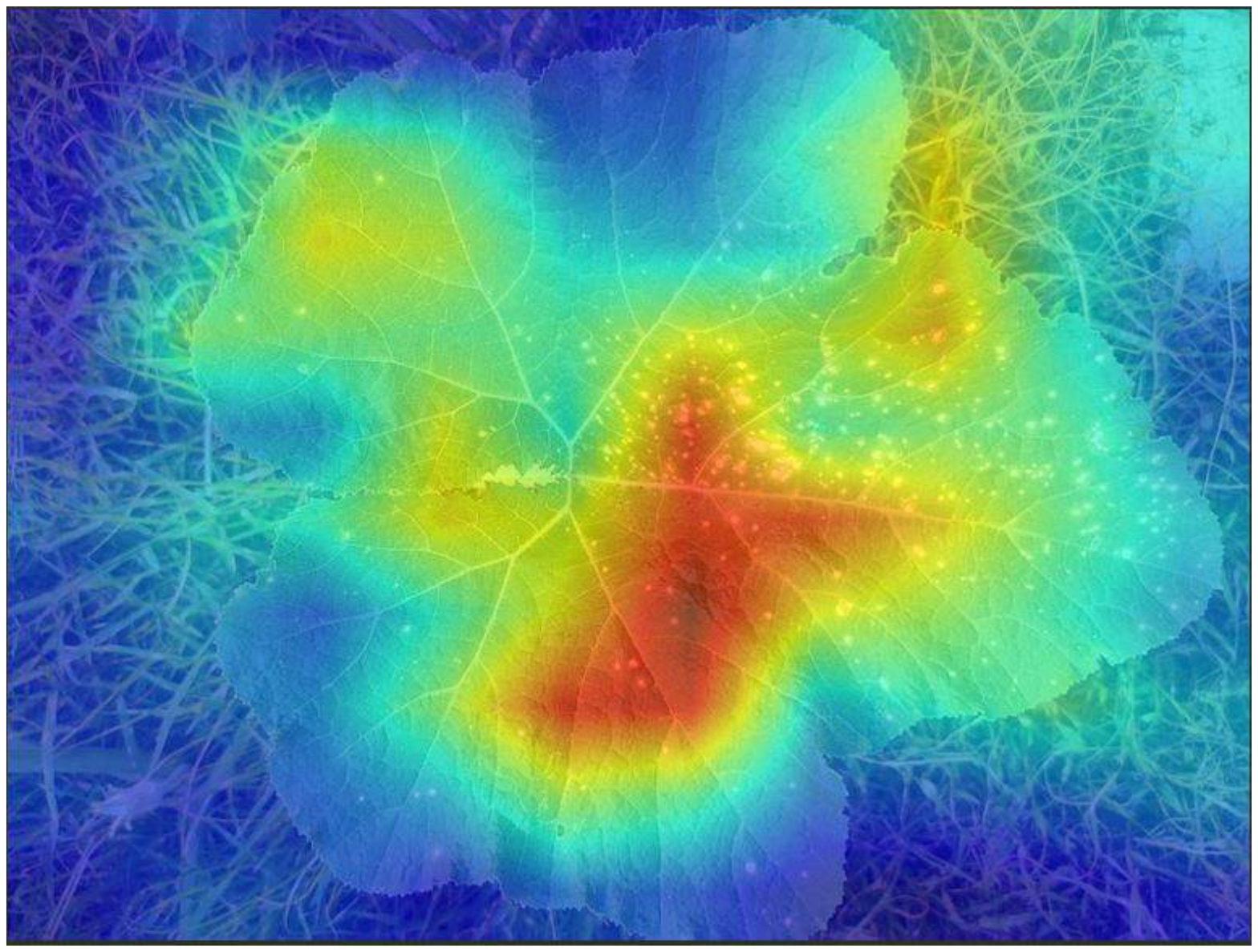}
        \caption{LayerCAM}
        \label{fig:sub3}
    \end{subfigure}
    \hfill
    \begin{subfigure}[b]{0.24\textwidth}
        \includegraphics[width=\textwidth]{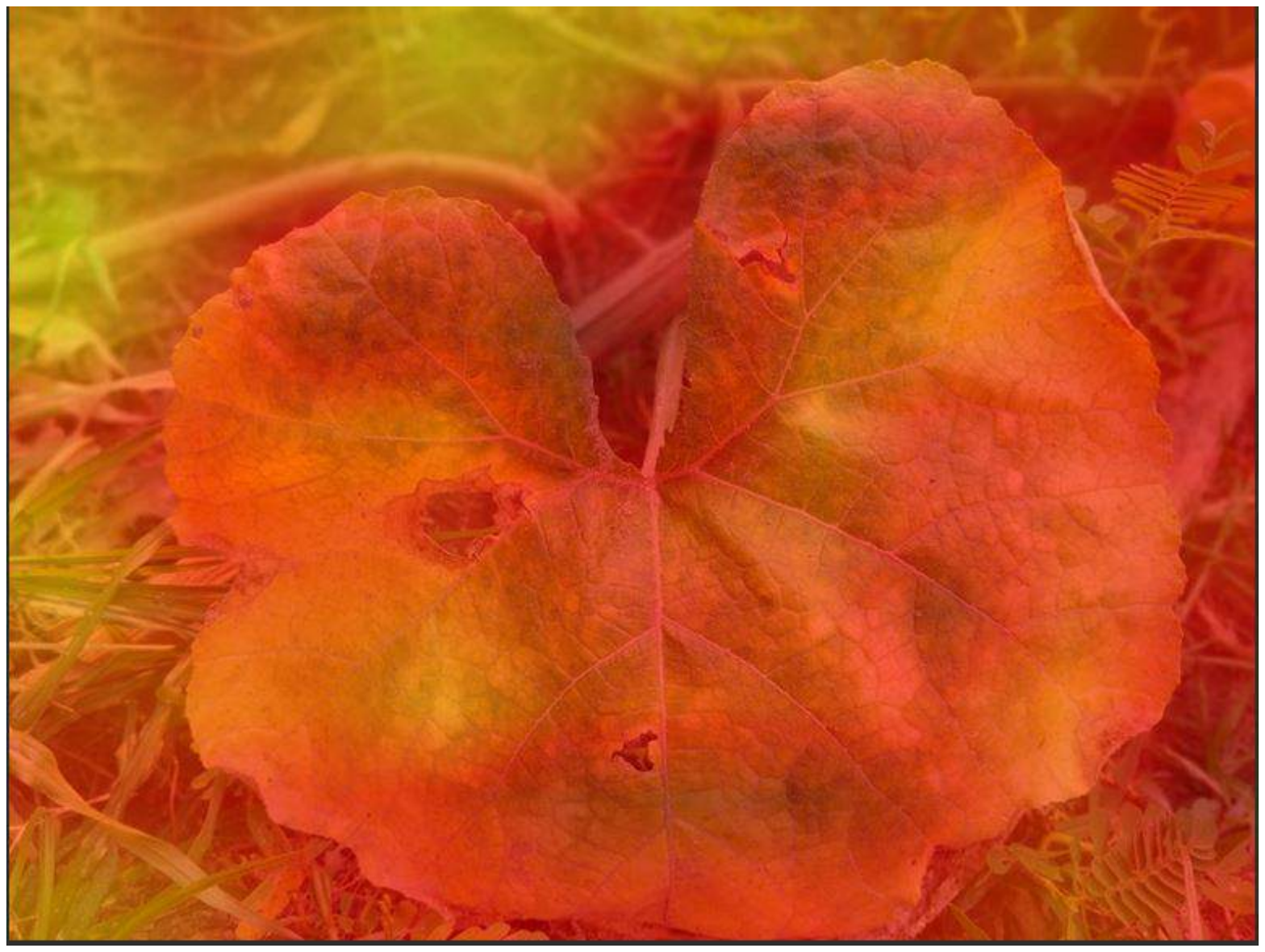}
        \caption{ScoreCAM}
        \label{fig:sub4}
    \end{subfigure}
    \caption{AI insights, Including GradCAM, GradCAM++, ScoreCAM, and LayerCAM, Provides Distinct Perspectives on the Source Image for Pumpkin Leaf Classification.}
    \label{fig:explanation_using_XAI}
\end{figure*}
The findings emphasize the importance of hyperparameter adjustments, as experiments across several configurations found that an epoch count of 50, a batch size of 10, and a learning rate of 1e-5 produced the best results for most models. ResNet50's excellent efficiency is explained by its efficient design, which balances depth and feature extraction, making it ideal for this fine-grained task. Deeper models, such as ResNet101 and DenseNet201, will likely be overfitted, resulting in lower generalization performance. While Xception and InceptionResNetV2 are sophisticated architectures, their lower performance on this dataset suggests that models with richer feature extraction capabilities, such as ResNet and DenseNet, are better suited for disease classification tasks using pumpkin leaf pictures. However, significant hyperparameter optimization was used to determine the best configuration for training deep learning models. The studies tested alternative learning rates (1e-3 and 1e-5) and epoch counts (30, 50, and 100) to see how they affected model performance as shown in TABLE \ref{tab1}. 

In these trials, it was discovered that a learning rate of 1e-5 produced greater convergence and generalization than 1e-3, which frequently resulted in overfitting or slower advances throughout training. Furthermore, the epoch count has a considerable impact in achieving ideal outcomes. While 30 epochs resulted in underfitting, especially for more complicated architectures, 50 epochs achieved a balance between training efficiency and generalization. On the other the side, increasing the number of epochs to 100 frequently resulted in overfitting, particularly for deeper models including ResNet101 and DenseNet201. After experimenting with several configurations, the combination of 50 epochs, a batch size of 10, and a learning rate of 1e-5 consistently generated the best results across all models, particularly ResNet50, which outperformed all other models. 
\balance
The confusion matrices presented demonstrate the performance of different deep-learning models in classifying leaf diseases. The ResNet50 model beats the others in terms of accuracy and consistency across various disease categories, including bacterial leaf spot, downy mildew, healthy leaf, mosaic disease, and powdery mildew. The ResNet50 matrix displays prevailing diagonal values indicating correctly identified diseases with minimal misclassifications when compared to other models such as DenseNet121, DenseNet169, DenseNet201, InceptionResNetV2, ResNet101, and Xception, illustrating its superiority in handling complex patterns in leaf disease detection. The finding exhibits ResNet50's robustness and reliability for actual agricultural diagnostics applications in Fig. \ref{fig:confusionmatrix}.

\subsection{Model Interpretation using XAI}
 In the present investigation, we've employed Explainable Artificial Intelligence (XAI) techniques—GradCAM, GradCAM++ , LayerCAM , and ScoreCAM to provide a visual representation of its decision-making process in detecting leaf illnesses. GradCAM and GradCAM++ emphasize significant areas of the images, increasing model transparency by revealing wherever the model focuses to generate predictions. LayerCAM provides more information by displaying activation across many layers, whereas ScoreCAM utilizes forward passing scores to build more consistent and representative activation maps. These strategies contribute to a better understanding of how the deep learning models processes visual data, ensuring that the model's diagnostic predictions are both trustworthy and interpretable.
 We have provided our best-performing model ResNet50's XAI images, as shown in Fig. \ref{fig:explanation_using_XAI}. These images illustrate how the model makes decisions, highlighting parts of the leaf images that have significant effects on the classification results. In addition to improving the model's interpretability, these visual explanations assist in identifying the patterns most suggestive of particular diseases. 
\section{Limitations \& Future Work}
\label{sec:limitation}
Though producing promising results, possesses limitations. The dataset contains 2,000 images, which may not fully represent the diversity of real-world farming panoramas. Moreover, our model selection was confined to CNN architectures, while deeper models such as ResNet101 and DenseNet201 demonstrated evidence of overfitting. In terms of XAI methodologies, we mostly used heat map visualizations, leaving other interpretability methods unexplored. In the future, increasing the dataset and investigating sophisticated deep learning models like as Vision Transformers and hybrid architectures could increase classification performance. Besides, using other XAI approaches, such as FasterScoreCam, SmoothGrad, lime, shap or Integrated Gradients, may improve the interpretability and utility of AI-driven diagnostics in agriculture. Although effective, the current approach can be further improved by enlarging the dataset to more accurately represent the variability in agricultural environments. Furthermore, the investigation of transfer learning techniques could be beneficial in reducing overfitting and enhancing the robustness of the model.

\section{Conclusion }
\label{sec:conclusion}
Our study introduces an effective approach for detecting pumpkin leaf diseases using advanced deep learning models combined with Explainable AI (XAI) techniques, improving both accuracy and interpretability. Among the tested models, ResNet50 demonstrated the highest accuracy at 90.5\%, while InceptionResNetV2 had the lowest accuracy at 81.5\%. Other models included DenseNet169 with an accuracy of 86.2\%, Xception at 88.5\%, ResNet101 at 87.5\%, DenseNet201 at 84.5\%, and DenseNet121 at 84.1\%. A key novelty of our study is the integration of XAI methods such as GradCAM, GradCAM++, ScoreCAM, and LayerCAM which provide clear visual insights into the features influencing the model's predictions. Unlike previous studies on pumpkin disease detection, which did not utilize XAI, our approach offers enhanced transparency and trust in AI-driven results. This improvement makes the outcomes more actionable for practical agricultural use. In conclusion, our study highlights the value of combining deep learning with XAI, providing a novel, interpretable solution for precise plant disease detection, ultimately aiding in better crop health management.


\vspace{12pt}

\end{document}